# Deep Causal Learning: Representation, Discovery and Inference

ZIZHEN DENG, XIAOLONG ZHENG*, School of Artificial Intelligence, University of Chinese Academy of Sciences; State Key Laboratory of Multimodal Artificial Intelligence Systems, Institute of Automation, Chinese Academy of Sciences, China

HU TIAN, Guanghua School of Management, Peking University, China

DANIEL DAJUN ZENG, School of Artificial Intelligence, University of Chinese Academy of Sciences; State Key Laboratory of Multimodal Artificial Intelligence Systems, Institute of Automation, Chinese Academy of Sciences, China

Causal learning has garnered significant attention in recent years because it reveals the essential relationships that underpin phenomena and delineates the mechanisms by which the world evolves. Nevertheless, traditional causal learning methods face numerous challenges and limitations, including high-dimensional, unstructured variables, combinatorial optimization problems, unobserved confounders, selection biases, and estimation inaccuracies. Deep causal learning, which leverages deep neural networks, offers innovative insights and solutions for addressing these challenges. Although numerous deep learning-based methods for causal discovery and inference have been proposed, there remains a dearth of reviews examining the underlying mechanisms by which deep learning can enhance causal learning. In this article, we comprehensively review how deep learning can contribute to causal learning by tackling traditional challenges across three key dimensions: representation, discovery, and inference. We emphasize that deep causal learning is pivotal for advancing the theoretical frontiers and broadening the practical applications of causal science. We conclude by summarizing open issues and outlining potential directions for future research.

## 1 INTRODUCTION

Causality has always been a very important part of scientific research [1-3]. It has been studied in many fields, such as biology [4-5], medicine [6-10], economics [11-15], epidemiology [16-18], and sociology [19-23]. For the construction of general artificial intelligence systems, causality is also indispensable [2, 24-25]. For a long time, causal discovery and causal inference have been the main research directions of causal learning. Causal discovery (CD) [26-27] is to find causal relationships from observational or intervention data, usually in the form of a causal graph. Causal inference (CI) [3, 28] is to estimate the causal effect between variables, which can be further divided into causal identification and causal estimation [1]. In causal identification, whether the causal effect can be estimated based on the existing information is determined, and in causal estimation, specific causal effect values are obtained.

Despite the existence of numerous methods in the field of causal learning, many problems remain unsolved. For causal discovery, most methods require strong assumptions, such as causal Markov conditions, and faithfulness assumptions [29-31], which are often impossible to verify. In addition, most traditional causal discovery methods are based on combinatorial optimization [27], a process that becomes comparatively intractable as the number of nodes increases. In causal inference, the absence of counterfactual data and the impracticality of randomized controlled trials (RCTs) often necessitate the use of observational data to estimate causal effects. The key problem with causal inference from observational data is selection bias [3], including confounding bias (from confounders) and data selection bias (from colliders). Selection bias can lead us to observe false causality or mistake correlation for causation. Traditional causal inference methods often have significant estimation bias due to limited fitting ability [32-33]. There are also several persistent issues in both causal discovery and causal inference, such as unknown interventions [34], unobserved confounders [35], and missing data [26]. In addition, causal discovery and inference have predominantly focused on structured data. Nevertheless, as application domains have broadened, there is a growing need to process unstructured data types, such as images, text, and video [36-40]. When dealing with unstructured data, the high-dimensional, low-level unstructured representations need to be transformed into low-dimensional, high-level structured causal representations.

We review the use of deep learning methods to address the aforementioned problems in the causal learning field. The three core strengths of deep learning for causal learning are strong **representational capabilities**, strong **fitting capabilities** and the ability to approximate **data generation mechanisms**. Firstly, when dealing with unstructured data, the representational power of neural networks ensures that they can obtain a suitable causal representation for causal discovery and inference [24, 41-42]. Secondly, due to their general fitting ability and flexibility of neural networks, they have become the primary method for continuous optimization to solve the long-standing combinatorial optimization problem in causal discovery, and they can theoretically cope with large-scale data. The universal approximation theorem indicates that neural networks can learn extremely complex functions for estimating heterogeneous treatment effects using low-bias estimators [43-44]. Thirdly, deep learning can explicitly or implicitly model data generation mechanisms. For instance, adversarial learning implicitly generates counterfactuals, commonly implemented with generative adversarial networks (GANs) [45], while disentanglement mechanisms that explicitly model the data generation process to generate proxy/latent variables, commonly implemented with variational autoencoders (VAEs) [46]. These advantages respectively solve or mitigate various challenges in causal learning, thus providing compelling reasons for integrating deep learning techniques into the field. In this article, we argue that causal representation learning, deep causal discovery, and deep causal inference together constitute the field of **deep causal learning**, as these three parts cover the general process of exploring causality: representation, discovery, and inference. Figure 1 shows the main differences between traditional causal learning and deep causal learning. We can clearly see the improvements that deep learning brings to causal learning.

In the past few years, many deep learning-based causal discovery and causal inference methods have been proposed. There have been many reviews related to causal learning, but few have summarized how to take advantage of deep learning techniques to improve causal learning methods. Guo et al. [25] reviewed causal discovery and causal inference methods with observational data, but very little of the content was related to deep learning. Yao et al. [28] focused on causal inference based on the potential outcome framework. It also mentioned some causal inference methods based on neural networks, but the description was not systematic. Nogueira et al. [47] summarized causal discovery and causal inference datasets, software, evaluation metrics and running examples without focusing on the theoretical level. Glymour et al. [26] mainly reviewed traditional causal discovery methods, and Vowels et al. [27] focused on continuous optimization discovery methods. Gong et al. [48] summarized the temporal causal discovery methods in-depth, but did not involve non-temporal data and causal inference. There have been reviews combining causality with machine learning [24, 49-50], but these survey papers have mainly explored how causality can be used to solve problems in the machine learning community. The work of Koch et al. [51] and Li et al. [52] are more similar to our starting point, and focused on the improvements that deep learning brings to causal learning. However, they only considered the combination of deep learning and causal inference and did not address other aspects of the field of causal learning, such as the representation of causal variables, causal discovery.

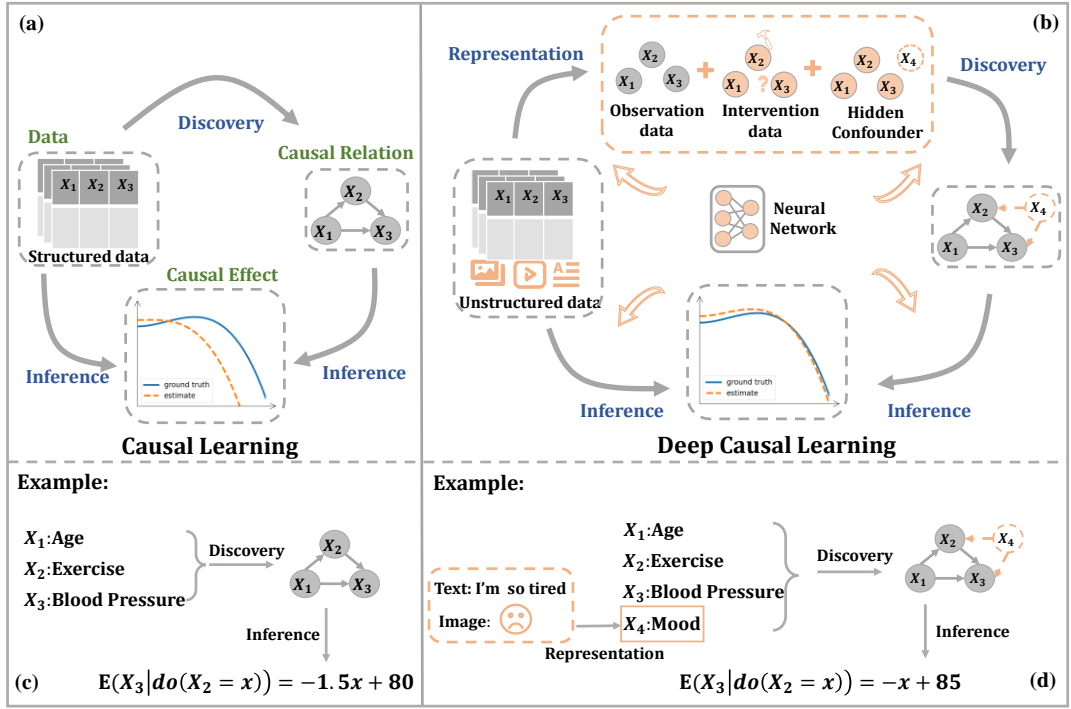

Figure 1: The difference between causal learning and deep causal learning. The comparison between (a) and (b) shows the theoretical advantages of deep causal learning. In the framework of deep causal learning, unstructured data can be processed with the representational power of neural networks. With the modeling capabilities of neural networks, in causal discovery, observational data and (known or unknown) intervention data can be comprehensively used in the presence of hidden confounders to obtain a causal graph that is closer to the facts. With the fitting ability of neural networks, the estimation bias of causal effects can be reduced in causal inference. The 4 orange arrows represent the neural network's empowerment of representation, discovery, and inference. (c) and (d) demonstrate the advantages of deep causal learning in more detail by exploring examples of the effect of exercise on blood pressure. We assume that the ground truth of exercise on blood pressure is $E(X_3 | do(X_2 = x)) = -1.1x + 84$.

We present a more comprehensive and detailed review of the changes that deep learning brings to causal learning. The rest of this article is organized as follows. Section 2 provides basic concepts related to causality. Section 3 reviews causal representation learning using observation and intervention data when dealing with unstructured data. Section 4 reviews deep causal discovery methods using i.i.d data and time series data. Section 5 reviews the deep causal inference methods based on covariate balance, adversarial training and proxy variables. Section 6 introduces new frontiers of causal learning. Section 7 provides a conclusion and discusses the future directions of deep causal learning. Figure 2 illustrates the overall framework of this survey.

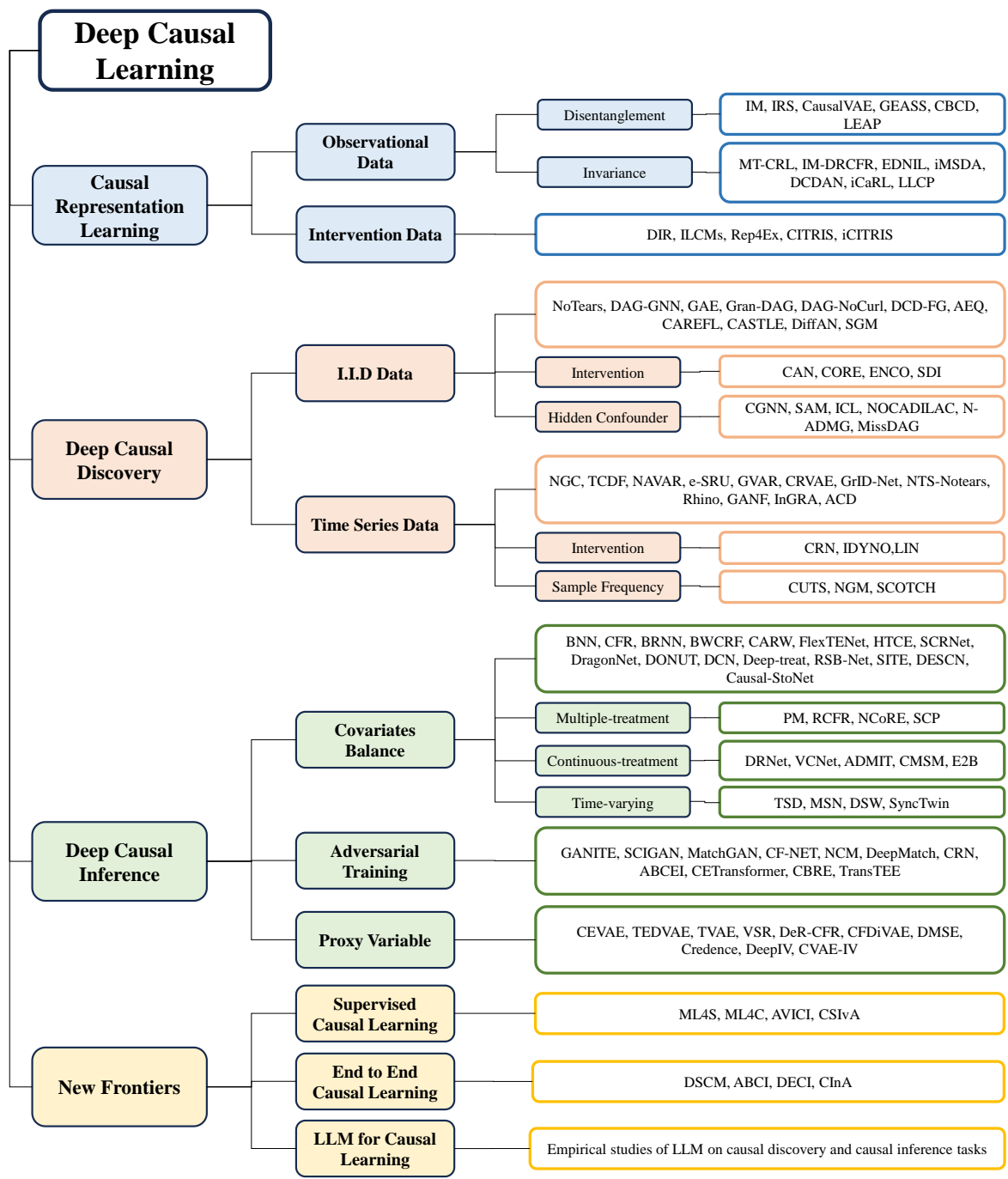

Figure 2: An overview of the main structure of the survey.

## 2 PRELIMINARIES

In this section, we briefly introduce the basic concept of causal discovery and causal inference. Table 1 presents the basic notations used in this article. Further comprehensive and detailed causal concepts and knowledge can be found in the literature [1, 3, 31, 53].

Table 1: Basic notations and their corresponding descriptions

| Notation | Description | Notation | Description |
|---|---|---|---|
| $X_i$ | Covariates of the sample $i$. | $V$ | The set of endogenous variables. |
| $T_i$ | Treatment of the sample $i$. | $U$ | The set of exogenous variables. |
| $Y_i^F$ | Factual outcome of the sample $i$. | $F$ | The set of mapping functions. |
| $Y_i^{CF}$ | Counterfactual outcome of the sample $i$. | $Pa_v$ | The parent nodes of $v$. |
| $e(X_i)$ | Propensity score of $X_i$. | $P(U)$ | Distribution of exogenous variables. |
| $\Phi(X_i)$ | Covariates representation of $X_i$. | $I$ | Instrumental variable. |
| $\theta$ | Neural network parameters. | $Z$ | Hidden/latent factors. |
| $h(\cdot)$ | Neural network mapping. | $dis(\cdot)$ | Distance measure. |
| $IPM_G(\cdot)$ | Integral probability metrics. | $E_p$ | The expectation based on data distribution $p$. |
| $\|\cdot\|_p$ | $p$-norm. | $N(\mu, \sigma^2)$ | Gaussian distribution with mean $\mu$ and variance $\sigma^2$. |
| $Loss(\cdot)$ | Loss function. | $\alpha, \beta, \gamma$ | Hyperparameters. |

### 2.1 Causal Model

*Structure Causal Model.* A structural causal model (SCM) is a 4-tuple $< V, U, F, P(U) >$, where $V$ is the set of endogenous variables, $U$ is the set of exogenous variables, $F$ is the set of mapping functions accomplish the mapping from the parent node $Pa_v$ to $v$, and $P(U)$ is the distribution of exogenous variables [3, 53].

In essence, SCM is a subjective abstraction of the objective world, and the involved endogenous and exogenous variables are heavily dependent on the researchers' prior knowledge. That is, the definitions of these variables themselves are not necessarily accurate, or the most essential variables cannot be observed due to various limitations. For example, when studying the effect of a person's family status on their academic performance, we might use the family's annual income as a proxy variable, although this variable may not be entirely appropriate or even correct.

Usually, each SCM model has a corresponding causal graph $G$, which is typically a directed acyclic graph (DAG) [3]. In fact, the casual graph can be seen as an integral part of SCM, in addition to counterfactual logic. As shown in Figure 3, there are three basic structures in a causal graph: Chain (a), Fork (b) and Collider (c). These three basic structures constitute a variety of causal graphs. In Figure 3 (e), $X$ represent covariates, $Y$ represent outcome, $T$ represent treatment, $C$ represent confounders, $I$ represent instrumental variable, $U_t$ represent the exogenous variable of $T$ and $U_y$ represent the exogenous variable of $Y$.

*d-separation.* In a causal graph $G$, we say that two sets of nodes $X$ and $Y$ are d-separated by a third set of nodes $\mathbf{Z}$, where $X, Y$ and $\mathbf{Z}$ are pairwise disjoint, if $\mathbf{Z}$ block all the path between nodes in $X$ and $Y$ [3].

*Intervention.* Simply put, intervening on node $X$ is to remove all edges pointing to $X$ on the causal graph $G$. The intervention distribution can be expressed as $P\big(Y\big|do(X = x)\big)$, which means assign $X = x$.

*Back-door adjustment.* In the causal graph corresponding to the SCM, there is a pair of ordered variables $(X, Y)$. If the variable set $Z$ satisfies the condition that there is no descendant node of $X$ in $Z$, and $Z$ blocks every path between $X$ and $Y$ pointing to the $X$ path, then $Z$ is said to satisfy the backdoor criterion [1] about $(X, Y)$, as shown in Figure 3 (d). If the variable set $Z$ satisfies the backdoor criterion of $(X, Y)$, then the causal effect of $X$ on $Y$ can be calculated using the following formula:

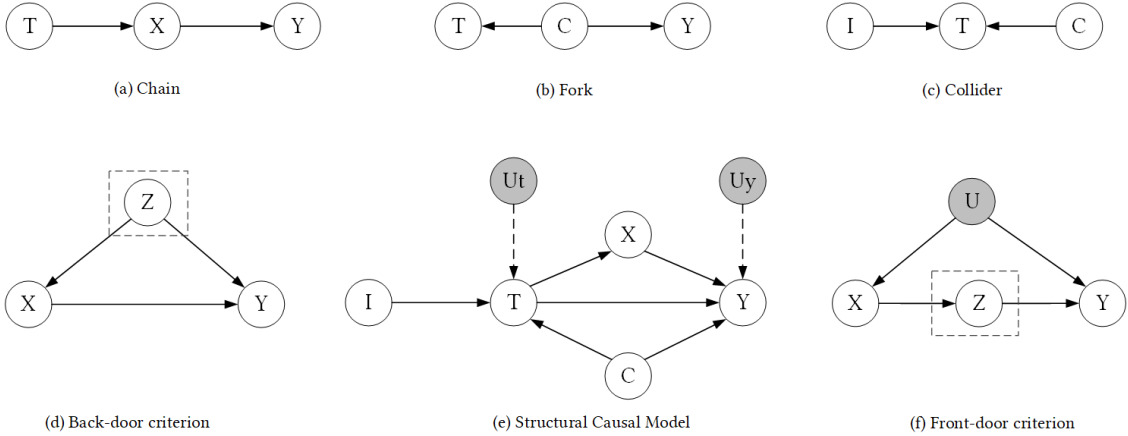

Figure 3: Three basic DAGs, a simple structure causal model, and two adjustment criteria.

$$P\big(Y=y\big|do(X=x)\big)=\sum_{z}P(Y=y|X=x,Z=z)P(Z=z)\,. \tag{1}$$

**Front-door adjustment.** As shown in Figure 3 (f), variable set $Z$ is said to satisfy the front door criterion [1] for an ordered variable pair $(X,Y)$, if the following conditions are satisfied: 1) $z$ cuts off all directed paths from $X$ to $Y$; 2) $X$ to $Z$ has no backdoor paths; and 3) all $Z$ backdoor paths to $Y$ are blocked by $X$. If $Z$ satisfies the front door criterion for the variable pair $(X,Y)$, and $P(x,z)>0$, then the causal effect of $X$ on $Y$ is identifiable and is calculated by:

$$P\big(Y=y\big|do(X=x)\big)=\sum_{z}P(z|x)\sum_{x'}P(y|x',z)P(x')\,. \tag{2}$$

**Causal sufficiency**. For a pair of observed variables $T$ and $Y$, all their common cause must be observed in the data, which means all common cause modeled in the causal graph $G$ [31].

## 2.2 Causal Discovery

Using observational data to discover causal relationships between variables is a fundamental and important problem. Causal relations among variables are usually described using a DAG, with the nodes representing variables and the edges indicating probabilistic relations among them. During causal discovery, the causal Markov conditions and faithfulness assumptions are often needed.

**Causal Markov condition**. In the graph $G$ of SCM, variables are conditionally independent given their cause variables [1].

**Faithfulness**. Any conditional independencies in the joint distribution are implied by the graph $G$, according to the d-separation properties [31].

Only when both the Causal Markov condition and the Faithfulness assumption are satisfied, the DAG $G$ can be regarded as a causal graph. The traditional causal discovery methods typically include constraint-based methods [31], score-based methods [54], function-based methods [55-56] and other classes of methods [57-59]. Constraint-based methods infer the causal graph by analyzing conditional independence in the data. Classical algorithms include PC and FCI [31]. The score-based methods search through the space of all possible DAGs representing the causal structure based on some form of scoring function for network structures. The classic scoring method for combinatorial optimization is GES [54], and the classic scoring method for continuous optimization is NoTears [60]. The function-based method requires assumptions about the data generation mechanism between variables and then the causal direction is judged through the asymmetry of the residuals. Essential algorithms include LiNGAM [55] and ANM [56]. For temporal causal discovery, in addition to the above methods, there are also methods based on Granger causality [11, 48, 61].

## 2.3 Causal Inference

To calculate various causal effects in the SCM framework, we must understand three forms of data corresponding to the Pearl's causal hierarchy (PCH) proposed by Judea Pearl [1, 62]: observation data, intervention data and counterfactual data. Observational data represent passive collection without any intervention. Causal effects cannot be calculated by relying solely on observational data without making any assumptions. Intervention refers to changing the value or distribution of a few variables, and the average treatment effect (ATE) may be calculated. Counterfactual data are unavailable in the real world. However, under various assumptions, individual treatment effects (ITE) can be calculated with the help of counterfactual theory. The capabilities of the three models were discussed in more detail in previous work [3].

Calculating causal effects is the core of causal inference. The average treatment effect (ATE) under the SCM framework is a common indicator to measure causal effects. It is defined as follows:

$$ATE = E[Y|do(t = 1)] - E[Y|do(t = 0)]. \tag{3}$$

As shown in Equation (3), the key to calculating the causal effect is to calculate the probability under the intervention. ATE may not capture differences in causal effects across subpopulations, which is overcome by the conditional average treatment effect (CATE):

$$CATE = E[Y|do(t = 1), X = x] - E[Y|do(t = 0), X = x]. \tag{4}$$

In the SCM framework, individual treatment effects (ITE) are generally not discussed, but in fact it is a special case of CATE. Here we borrow the POM (Potential Outcome Framework) [7, 18] to express the individual treatment effects (SCM and POM are equivalent in principle, but only from a different perspective [3]):

$$ITE = Y_i(t = 1) - Y_i(t = 0) \tag{5}$$

## 3 CAUSAL REPRESENTATION LEARNING

Causal discovery and causal inference always adhering to a basic assumption: the causal variables under study should be identified and available at the beginning of the analysis. However, sometimes we cannot directly obtain low-dimensional, high-level, meaningful causal variables, but high-dimensional, low-level non-causal variables, such as words in text, pixels in images and videos. Therefore, we need to transform unstructured variables into structured variables for causal discovery (CD) and causal inference (CI), a process known as Causal Representation Learning (CRL). In other words, CRL is not a necessary prerequisite for CD and CI, but a countermeasure when dealing with unstructured data.

In fact, the original intention behind developing CRL is to deal with the generalization and interpretability problems in representation learning [24]. Variables' causal representations are obtained as intermediate products in this process, enabling the handling of unstructured data issues in causal discovery and causal inference. This section discusses only those causal representation learning methods that explicitly generate causal representations.

Generally speaking, in order to identify causal representations, the distribution of the data must be altered to find an invariant model within it that can determine causality. Based on the source of the distribution change, we categorize the existing methods into two types: observational data, where domain changes cause distribution shifts, and intervention data, where interventions cause distribution shifts.

## 3.1 Causal Representation Learning using Observational data

The starting point of most causal representation learning is to obtain causal representations, thereby improving the generalization and interpretability of the model, supporting intervention, reasoning, and planning. Many studies have achieved this goal from different perspectives. This section will introduce how different methods can obtain causal representation from the perspectives of disentanglement and invariance.

***Disentanglement perspective.*** IM [63] is a modular system that autonomously discerns independent mechanisms from unlabeled, shifted data. It can be adapted to various scenarios without prior knowledge of the number of underlying mechanisms. Using a multi-expert approach, it leverages competition among smaller neural networks to uncover causal models. During training, only the winning expert updates its weights, while others remain static. IRS [64] introduces a robustness metric for assessing disentangled representation learning, viewing disentanglement as a causal process rather than a mere heuristic feature in data generation. CausalVAE [65] assumes linear causality, utilizing a causal layer to transform independent exogenous variables into causally significant endogenous ones. It encodes observed variables to generate exogenous inputs, which are then processed by the causal layer for endogenous causal impact. GEASS [66] employs a neural network to detect sparse features with Granger causality, ensuring causal feature recovery through maximizing modified transfer entropy with sparsity. This approach uses a stochastic gate layer for sparse feature subset selection, applicable to high-dimensional data. CBCD [67] uses VAE to derive binary causal variables from unstructured data, explaining classifier outcomes. It's a partial causal discovery technique, focusing on identifying causal variables without exploring their interrelationships. Yao et al. [68] developed a nonparametric causal process identifiability theory, exploring how distribution shifts can enhance decoupling. The framework assumes reversible nonlinear mixing and decomposes distribution shifts under both constant and variable causal relationships. It recovers time-delayed causal variables and identifies their relationships from time series data in stationary environments with distribution shifts. LEAP [69] targets the recovery and relationship determination of latent causal variables with time lags from time series data. It introduces two verifiable conditions for temporal causal process identification from nonlinear mixtures and lays a theoretical groundwork for estimating latent temporal causality.

***Invariance perspective.*** MT-CRL [70] seeks to capture multi-task knowledge with decoupled neural modules and robust task-routing through invariance-based regularization. The modules, realized via a multi-gated Mixture-of-Experts, focus on distinct data aspects. A dual-layer optimization approach is employed: the outer layer refines the encoder and routing, while the inner layer enhances task-specific predictors. This enables shared representation learning alongside task-specific accuracy. IM-DRCFR [71] employs mutual information minimization for learning counterfactual regression of disentangled representations, ensuring their independence. EDNIL [72] features a dual-module multi-head network, with an environment inference model to infer labels and mitigate bias, and an invariant learning model for identifying cross-environment invariant features. Kong et al. [73] introduced iMSDA, a domain adaptation framework using a variational autoencoder to learn separable representations with an invariant component serving as a causal representation. DCDAN [74] acquires domain-invariant causal representations by simulating target domain samples and reweighting source samples to focus on causal effects over correlations. iCaRL [75] builds invariant predictors by identifying latent variables with NF-iVAE (non-factored identifiable variational autoencoder), discovering causal graphs, and learning based on these direct causes within a diverse training setup.

Multimodal data involves concurrent observations from related sources like images and their captions. Imant et al. [76] explored identifiability in multimodal contrastive learning, revealing that it can block-identify shared latent factors across modalities without pinpointing individuals. LLCP [77], a temporal multivariate model, leverages causal modularity for local processes, enabling root cause identification and counterfactual outcome prediction. It comprises an encoder for latent space mapping, a decoder for variable generation, and a module for causal dynamics modeling.

### 3.2 Causal Representation Learning using Intervention data

Compared with the natural distribution shift caused by domain changes in observed data, the information about distribution changes induced by intervention is clearer and more conducive to identifying invariant causal representations.

DIR [78] generates interventional distributions to identify stable causal rationales, separating causal from non-causal subgraphs and intervening to discern invariant causal elements. ILCMs [79] employ a weakly supervised method with unlabeled paired samples, utilizing variational autoencoders to model causal variables and structures without explicit graph optimization. Sorawit et al.'s Rep4Ex [80] leverages autoencoders for invariant latent mappings and control functions to estimate intervention effects, capturing latent causal representations.

CITRIS [81] targets causal representation learning from image time series potentially influenced by multidimensional causal factors, such as 3D object interactions. It employs normalization flows to disentangle representations from a pretrained autoencoder. iCITRIS [82] extends this by accommodating immediate effects in observable interventions, identifying latent multidimensional causal variables and their graphs through differential causal discovery methods. It's built on a variational autoencoder framework with convolutional encoders and decoders for reversible data mapping, enabling distinct latent space representations for each causal variable.

## 4 DEEP CAUSAL DISCOVERY

This section introduces causal discovery methods based on deep neural networks. Traditional causal discovery methods are combinatorial optimization problems, and the technique commonly used in training NNs is continuous optimization [83-84]. In recent years, numerous methods using neural networks for causal discovery have emerged, and the field of causal discovery has been expanded and supplemented from different angles. Generally speaking, deep causal discovery can be discussed from two perspectives: independent and identically distributed data (i.i.d.) and time series data.

### 4.1 Causal Discovery from I.I.D. Data

This subsection introduces methods for causal discovery on observational i.i.d. data using deep neural networks. These methods usually convert the constraints of acyclicity into constraints on the trace of the adjacency matrix. NOTEARS [60] was the first to transform causal discovery from a combinatorial optimization problem to a continuous optimization problem, setting the stage for the subsequent introduction of neural networks. Specifically, the previous causal discovery methods usually require the obtained causal graph to be a DAG. However, with the increase in the number of nodes, the complexity of the combinatorial optimization problem increases very rapidly, and the speed of solving the problem is greatly reduced, which seriously limits the solvability size of the problem. The transformation of the form is as follows. In Equation (6), $A$ represents the adjacency matrix of the causal graph, $d$ represents the number of nodes, and $h(\cdot)$ is a smooth function over real matrices. This derivation is concise and powerful but also ingenious. Most of the subsequent gradient-based methods are extensions of this. Note that in the formula, the value of $h(A)$ is usually small but not 0, so the setting of the threshold is required in most cases.

$$G(A) \in DAGs \Leftrightarrow h(A) = 0 \Leftrightarrow tr\left(e^{A^\circ A}\right) - d = 0. \tag{6}$$

NOTEARS has a good performance under the assumption of the SEM, and many subsequent works have extended it to the nonlinear field. DAG-GNN [85] integrates neural networks into causal discovery for nonlinear scenarios, using encoder-decoder architecture to derive adjacency matrices. Following the idea of encoder-decoder, GAE [86] utilizes the idea of graph self-encoder to better utilize graph structure information for causal graph structure discovery. Another method that uses neural networks to adapt to nonlinear scenarios is GraN-DAG [87]. It can handle the parameter families of various conditional probability distributions. The idea behind it is similar to that of DAG-GNN, but it achieves better results in experiments. DAG-NoCurl [88] combines GNN with the augmented Lagrangian method for nonlinear SEMs. DCD-FG [89] employs a Gaussian nonlinear low-rank SEM model with MLPs for factor nodes, using Gumbel-softmax sampling to prevent self-loops. A new indicator based on the reconstruction error of the autoencoder is proposed in AEQ [90]. Different indicator values are used to distinguish the causal directions, and the identification effect is better under univariate conditions. CASTLE [91] uses the adjacency matrix of the DAG is learned in the process of continuous optimization and embeds it into the input layer of the FNN.

At the same time, there are a large number of deep generative models used for causal discovery, modeling complex relationships between variables. CAREFL [92] delves into the link between normalizing flows and causality, connecting autoregressive flows with identifiable causal models and introducing a likelihood-based causal direction metric for multivariate data. SGM [93] performs generative modeling through data distribution gradient estimation using standard ReLU neural networks. DiffAN [94] employs diffusion probability models to learn score functions for causal graph topological sorting, offering efficiency over DAG space searches and scalability for large datasets with many variables and samples.

***Intervention data.*** Intervention data reveal more information about data-generating mechanisms and allow us to obtain a more accurate causal graph. CAN [95] generates diverse samples from conditional and interventional distributions, using mask vectors for intervention selection. However, it does not guarantee finding true causal graphs due to potential data assumption mismatches. CORE [96] treats causal discovery as a Markov decision process, using a Deep Q Network for structural updates and interventions in a state space of causal models. ENCO [97] sidesteps acyclicity constraints by optimizing independent edge likelihoods as an edge direction parameter problem, scaling to larger graphs with gradient estimators. It employs neural networks to model conditional distributions, scores graphs on intervention data, and optimizes graph parameters with an objective function combining log-likelihood and edge probability regularization.

Interventions may also be unknown, which means not known which variables were intervened. A new class of approaches is needed to deal with this situation. SDI [34] facilitates concurrent discovery of causal diagrams and structural equations amidst unknown interventions. It acknowledges the interdependence of structural and functional parameters, training the DAG's structural representation and independent causal mechanisms' functional representation in tandem. The method begins with parameterization, distinguishing between structural (adjacency matrix) and functional (M functions) parameters. A multilayer perceptron refines these parameters against observed data. The graph score, derived from intervention data, incorporates a cyclicity penalty.

***Hidden confounders.*** In practice, we do not necessarily have access to all variables because not all variables contribute to causal discovery. However, when confounding variables are not observed, causal discovery will face various challenges. One of the most serious problems is that the unobserved confounders can lead to an observed association between two variables that do not have a causal relationship between them. In traditional methods, there are some methods to deal with unobserved confounders, such as FCI. However, most of them are based on combinatorial optimization [98], which is not efficient enough. Here, we introduce some causal discovery methods based on deep neural networks, which can more efficiently and accurately find causal relationships in the presence of unobserved confounders. In CGNN [99], the prior form of the function is not set and MMD is used as a metric to calculate the score of each graph to evaluate how well each causal graph fits the observed data. The most important contribution of this method is the formal definition of a functional causal model with latent variables; an exogenous variable $U_{ij}$ that affects variables $i$ and $j$ is set to solve with the presence of unobserved confounders and proved that it is still possible to learn with backpropagation. The maximum mean discrepancy (MMD) is defined as shown in Equation (7), and $k(\cdot)$ is the Gaussian kernel:

$$\widehat{MMD}_k(D, \widehat{D}) = \frac{1}{n^2} \sum_{i,j=1}^{n} k(x_i, x_j) + \frac{1}{n^2} \sum_{i,j=1}^{n} k(\hat{x}_i, \hat{x}_j) - \frac{2}{n^2} \sum_{i,j=1}^{n} k(x_i, \hat{x}_j). \tag{7}$$

SAM [100] solves the computational limitation of CGNN. Its equation differs from structural equations in that it incorporates all variables other than itself, hence it is termed "structural agnostic." In SAM, noise matrices instead of the noise variable of variable pairs. The introduction of differentiable parameters in the correlation matrix allows the SAM structure to efficiently and automatically exploit the relation between variables, thus providing SAM with a way to discover correlations from unobserved variables.

$$X_j = \hat{f}_j(X, U_j) = \sum_{k=1}^{n_h} m_{j,k} \emptyset_{j,k}(X, U_j) z_{j,k} + m_{j,0}, \tag{8}$$

where $\hat{f}_j$ is the nonlinear function, $\emptyset_{j,k}$ is the feature, $z_{j,k}$ is the Boolean vector and $E_j$ is the noise variable.

ICL [101] addresses hidden variables by iteratively imputing missing data, utilizing GAN for interpolation and VAE for causal structure learning to uncover joint distributions and causality. NOCADILAC [102] employs a nonlinear causal model with observed and unobserved variables, distinguishing between direct causal effects and confounder influences, using VAE for parameter learning and causal graph construction. N-ADMG [103] identifies potential confounding under acyclic directed mixed graph assumptions and nonlinear noise models, establishing ADMG identifiability conditions and proposing an autoregressive flow-based neural model for learning. MissDAG [104], grounded in the EM framework, maximizes observed data likelihood for causal graph and parameter estimation, integrating diverse causal discovery methods and adeptly managing various ANMs. Table 2 shows all the mentioned deep causal discovery methods from i.i.d data.

## 4.2 Causal Discovery from Time Series Data

Causal discovery in time series is the key to many fields in science [105]. In addition to common problems such as hidden confounders, time series data has its own unique challenges, such as non-stationary state, instantaneous causal effect, etc. [48, 106-107]. There are three types of causal graphs for time series: full time graphs, window causal graphs and summary causal graphs. Full time graphs depict the causal relationship between variables at each moment, window causal graph is simplified in units of maximum delay and summary causal graphs depict the causal relationship during this time. Therefore, the summary causal graph obtained from the time series is likely to have cycles, which does not satisfy the DAG condition.

Table 2: Deep causal discovery methods from i.i.d data.

| Method | Year | NNs | Hid. Con. | Inter. | Output |
|--------|------|-----|-----------|--------|--------|
| CGNN [99] | 2017 | VAE | Y | Y | DAG |
| DAG-GNN [85] | 2019 | VAE, GNN | N | N | DAG |
| GAE [86] | 2019 | GAE | N | N | DAG |
| Gran-DAG [87] | 2019 | MLP | N | N | DAG |
| SDI [34] | 2019 | MLP | N | Y | DAG |
| SAM [100] | 2019 | MLP | Y | N | DAG |
| AEQ [90] | 2020 | AE | N | N | Pair |
| CASTLE [91] | 2020 | AE | N | N | DAG |
| CAREFL [92] | 2020 | MLP | N | Y | DAG |
| ICL [101] | 2020 | GAN, VAE | Y | Y | DAG |
| CAN [95] | 2020 | GAN | N | Y | DAG |
| DAG-NoCurl [88] | 2021 | GNN, MLP | N | N | DAG |
| ENCO [97] | 2021 | MLP | N | Y | DAG |
| DiffAN [94] | 2022 | MLP | N | N | DAG |
| DCD-FG [89] | 2022 | MLP | N | Y | f-DAG |
| MissDAG [104] | 2022 | MLP | Y | N | DAG |
| SGM [93] | 2023 | MLP | N | N | DAG |
| N-ADMG [103] | 2023 | NAF | Y | Y | ADMG |
| NOCADILAC [102] | 2023 | VAE | Y | N | DAG |
| CORE [96] | 2024 | DQN, MLP | N | Y | DAG |

Table 2. 'Hid. Con.' indicates if there are hidden confounders. 'Inter.' indicates if there are intervention data.

The most common method for discovering causal relationships in time series is the Granger causal (GC) [108]. Although the GC can achieve common-sense results under linear assumptions, the results of the GC in nonlinear scenarios are often unsatisfactory. There are many variations of GC-based methods that address the nonlinear problem [61, 109-110]. NGC [61] separates the functional representation of each variable to achieve an effective distinction between cause and effect to a certain extent. NGC provides a neural network for each variable $i$ to calculate the influence of other variables on it. If a column of the obtained weight matrix is 0, it means that the corresponding variable has no Granger causality to the variable $i$. The core of the NGC is to design a structured sparsity-induced penalty to achieve regularization so that the Granger causality and the corresponding time delay $t$ can be selected at the same time:

$$min_W \sum_{t=K}^{T} \left( x_{it} - g_i \left( x_{(t-1):(t-K)} \right) \right)^2 + \lambda \sum_{j=1}^{P} \Omega(W_{:j}^1),$$ (9)

where $\Omega(\cdot)$ is the penalty, W is the weights matrix and $g_i$ is the function of the relationships among variables.

In addition to NGC, there are many other deep temporal causal discovery methods based on Granger causality. TCDF [111] uses an attention-based convolutional neural network combined with a causal verification step to learn the causal graph structure. The architecture of TCDF includes multiple independent attention-based CNNs, each of which uses the same architecture but has different target time series. By interpreting the internal parameters of the convolutional network, it is able to discover the time delay between the cause and the effect. NAVAR [112] employs an additive framework emphasizing the unique contributions of each input variable to outputs, quantifying and ranking their impacts through independent nonlinear functions. e-SRU [113] advances time series forecasting with a component-based approach using Statistical Recurrent Units for nonlinear dynamics, curbing overfitting by minimizing trainable parameters and employing random projections for efficient recursive computations. GVAR [114] presents a self-interpretable neural network framework for Generalised Vector Autoregression, analyzing both the direction and temporal dynamics of GC effects under nonlinear influences. CRVAE [115] features an encoder and multi-head decoder, with each head generating a time series dimension. It promotes sparse weight matrices to encode Granger causality through a sparsity penalty, and generates time series by sampling independent noise sets and iterating through the decoder. GrID-Net [116] enhances GNN architecture for Granger causal inference in single-cell data, enabling lagged message passing on DAGs to accumulate past information and proposing a neural network-tailored formula for recovering nonlinear long-term dependencies.

In addition to the ideas based on Granger causality, the scoring-based method is also closely combined with neural networks. NTS-NOTEARS [117] captures nonlinear and time-lagged relationships with one-dimensional convolutional neural networks (1D CNNs), training separate networks to predict target variables from their temporal context, and deriving graph structures from the first layer's kernel weights. Rhino [118] combines vector autoregression, deep learning, and variational inference to model nonlinear relationships with immediate effects and flexible history-dependent noise. It is able to capture nonlinear relationships between variables as well as immediate effects due to slow sampling intervals. GANF [119] introduces a flow model for causality, using Bayesian networks to decompose joint probabilities into conditional probabilities, with DAG and flow parameters estimated jointly.

In many real-world systems, we often encounter a large amount of multivariate time series data collected from different individuals, which share some commonalities but also have heterogeneity. Existing methods usually train a unique model for each individual, which faces the problems of inefficiency and overfitting. InGRA [120] analyzes multivariate time series data from various individuals to identify common causal patterns and predict future values for each person's target variables, while reconstructing unique causal structures for each. Similar ideas, ACD [121] focuses on the notion that different causal graphs within the same dynamical system share common information, seeking a model for causal discovery across such graphs. It samples from multiple graphs of the same system, treating each as a training instance, and extends the amortized encoder to predict additional variables and account for unobserved confounders using structural knowledge. By extending the amortized encoder, it is possible to predict an additional variable, combined with structural knowledge, which can be used to represent unobserved confounders.

*Intervention data*. CRN [122] tackles causal discovery in dynamic systems by training a learner to sample from varied causal mechanisms, synthesizing intervention data distributions to infer causal graphs. This approach facilitates domain adaptation and accumulates knowledge for structural learning. Each episode begins with a new graph target, random interventions, and neural network training on the adjacency matrix, enabling structural learning through diverse interventions. IDYNO [123] operates within a continuous optimization framework with acyclicity constraints, suitable for both observational and interventional time series. It extends DYNOTEARS, supports nonlinear objectives via neural models, and features a modified function for varied intervention target distributions. LIN [124] clusters data into domains, each representing a potential intervention area, using neural networks to parameterize variable conditional distributions within clusters. Parameters are derived from parent node values, and a lower triangular matrix captures conditional distribution similarities across clusters, ensuring identical distributions share the same matrix row values.

*Sampling frequency and continuous data.* In the real world, time series data often have random missing or non-uniform sampling frequency due to sensor limitations or transmission losses. Apart from that, the observation of time series data is usually continuous, but most existing identifiability results and learning algorithms assume that the underlying dynamics are discrete-time, and few studies explicitly define dependencies in infinitesimal intervals of time, which poses a challenge to existing causal discovery methods. CUTS [125] focuses on causal discovery from irregular time series, employing a neural Granger causal algorithm within an iterative framework to perform data interpolation and graph construction. It features two complementary modules: the latent data prediction module, which forecasts and registers high-dimensional, complexly distributed irregular data, and the causal graph fitting module, which constructs a sparse causal adjacency matrix using the interpolated data. NGM [126] proposes a score-based graphical model learning algorithm for learning dynamic systems, and proves that for a class of vector fields parameterized by large neural networks, a directed graph of local independence in a system of stochastic differential equations (SDE) can be consistently recovered through least squares optimization and an adaptive regularization scheme. SCOTCH [127] uses continuous-time SDE to simulate the dynamic changes of data, which enables it to more accurately capture the inherent randomness of the system, and it can handle irregularly sampled time series data and simulate systems with continuous processes. Table 3 shows all the mentioned deep causal discovery methods from time series data.

Table 3: Deep causal discovery methods from time series data.

| Method | Year | NNs | Hid. Con. | Inter. | Instan. | Output graph |
|---|---|---|---|---|---|---|
| NGC [61] | 2018 | MLP, LSTM | N | N | N | Summary |
| TCDF [111] | 2019 | CNN, Attention | Y | N | N | Window |
| NAVAR [112] | 2020 | MLP, LSTM | N | N | N | Summary |
| e-SRU [113] | 2020 | SRU | N | N | Y | Summary |
| ACD [128] | 2020 | GNN | Y | N | N | Summary |
| CRN [122] | 2020 | LSTM | N | N | Y | Summary |
| GVAR [114] | 2021 | SENN | N | N | Y | Summary |
| InGRA [120] | 2021 | LSTM | N | N | N | Summary |
| NTS-NOTEARS [117] | 2021 | CNN | N | N | Y | Window |
| GANF [119] | 2022 | GANF | N | N | Y | Summary |
| GrID-Net [116] | 2022 | GNN | N | N | N | Window |
| IDYNO [123] | 2022 | MLP | N | Y | Y | Window |
| NGM [126] | 2022 | CNN | N | N | N | Summary |
| Rhino [118] | 2023 | MLP | N | N | Y | Window |
| CRVAE [115] | 2023 | RNN, VAE | N | N | Y | Summary |
| LIN [124] | 2023 | MLP | N | Y | N | Summary |
| CUTS [125] | 2023 | DSGNN | N | N | N | Summary |
| SCOTCH [127] | 2024 | Diffusion | N | N | Y | Window |

Table 3. 'Hid. Con.' indicates if there are hidden confounders. 'Inter.' indicates if there are intervention data. 'Instan.' indicates if there have instantaneous causal effect. 'Output graph' indicates whether the output is a window causal graph or a summary causal graph.

## 5  DEEP CAUSAL INFERENCE

Most traditional causal inference methods are applied to the original low-dimensional feature space. When the interactions among covariates are complex, this can result in substantial errors in estimating causal effects. As deep learning has gained popularity, numerous studies have started leveraging the robust fitting capabilities of neural networks to investigate the relationships between treatments and outcomes.

The core problems of causal inference are the missing counterfactual data and selection bias, as shown in Figure 4. The former is the more fundamental issues of the two, since once counterfactual data are available, the estimation of causal effects becomes very simple and straightforward. Therefore, from the perspective of solving the fundamental problems, the methods of causal inference can be divided into selection bias-oriented and counterfactual data-oriented methods. Further, existing deep learning-based causal inference methods can be roughly divided into three categories: covariate balance-based methods adjust covariates to balance the distribution of covariates in different treatment groups, thereby eliminating selection bias; adversarial training-based methods utilize adversarial training to make the discriminator unable to distinguish between the real data and the data generated by the generator, thereby realizing the generation of implicit counterfactual data; proxy variable-based methods model the data generation mechanism as the joint action between multiple latent variables to achieve explicit counterfactual generation. The relationship between the two core problems and three main methods is shown in Figure 4.

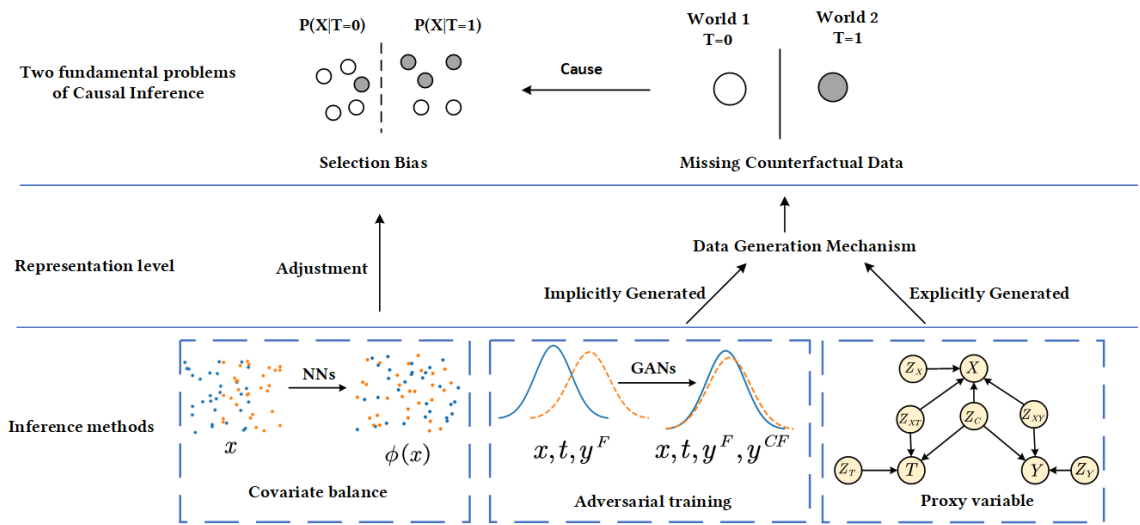

Figure 4: The methods to solve the fundamental problems of causal inference using deep learning. The most fundamental problem in causal inference is the missing counterfactual data. Due to the lack of counterfactual data, only observational data can be used to estimate causal effects, leading to selection bias. In essence, if the data generation mechanism can be modeled, then the "counterfactual data" can be approximated, and the problem of causal inference can be solved. In "Selection Bias", gray and white nodes represent individuals with different covariates. In "Missing Counterfactual Data", white and gray nodes represent outcomes under two treatments, i.e., fact and counterfactual, and only one of them can be observed.

## 5.1 Covariate Balance-based Causal Inference

In this subsection, we focus on causal inference methods based on the covariate balance perspective. The core of traditional methods for causal inference is to balance the covariates to estimate causal effects. Neural networks have a strong fitting ability so that through much training, the quantitative relationship between treatment $T$ and effect $Y$ can be found. This is also due to the strong fitting ability of the neural network; there must be regular terms to balance the covariates $X$ when used to estimate causal effects. Using neural networks, the counterfactual result ($y_{t=1}^{CF}$) of the control group ($t = 0$) with a certain value of the covariate ($X = x$) can be predicted, thus obtaining causal effects $y_{t=1}^{CF} - y_{t=0}^{F}$. Such methods typically use neural networks to obtain the representations of covariates or the propensity scores and then train estimators by minimizing the differences in representations between treatment and control groups. In the loss function, these balance ideas are usually achieved through regularization terms. The advantage of the neural network is that it can flexibly use various forms of regularization to balance the distribution of covariates to eliminate the influence of confounders. At the same time, it has a strong estimation ability and high accuracy. The general architecture of covariates balance-based causal inference methods is usually as shown in Figure 5.

The first method to use neural networks to achieve counterfactual predictions was BNN [129], which considers the problem of counterfactual inference from a domain adaptation perspective. The neural network is used to learn the representation of covariables, and then match is implemented in the representation space, i.e., selecting the nearest observation effect as its counterfactual effect for training. For example, for a sample $x_i$ in the treatment group ($t = 1$), the observation effect of sample $x_j$ nearest to the covariate representation in the control group ($t = 0$) is chosen as its counterfactual effect $y_i^{CF} = y_{j(i)}^{F}$. During training, each sample has two estimate outcomes $h(\Phi(x_i), t_i)$ and $h(\Phi(x_i), 1 - t_i)$. In Equation (10), the first term and third term represent the difference between the observed outcome and the predicted outcome. The discrepancy distance $dis_H$ is used to measure the difference between the two distributions. BNN was also the earliest causal representation learning method to balance latent space. The important contribution of the method is making a trade-off between balance and accuracy. The optimization objective is as follows:

$$B_{H,\alpha,\gamma}(\Phi,h) = \frac{1}{n}\sum_{i=1}^{n}\left|h(\Phi(x_i),t_i) - y_i^F\right| + \alpha\, dis_H\left(\hat{P}_\Phi^F, \hat{P}_\Phi^{CF}\right) + \frac{\gamma}{n}\sum_{i=1}^{n}\left|h(\Phi(x_i), 1-t_i) - y_{j(i)}^F\right|, \quad (10)$$

where $\Phi$ is the learned representation and $dis_H(\cdot,\cdot)$ is the distance measure. By minimizing the loss function equation (10), the BNN can simultaneously accomplish counterfactual inference and covariate balance. In the BNN, the network structure only has one head. This means that the treatment assignment information $t_i$ needs to be concatenated to the representation of covariate $\Phi(x_i)$. In most cases, $\Phi(x_i)$ is high-dimensional, so the information of $t_i$ might be lost during training.

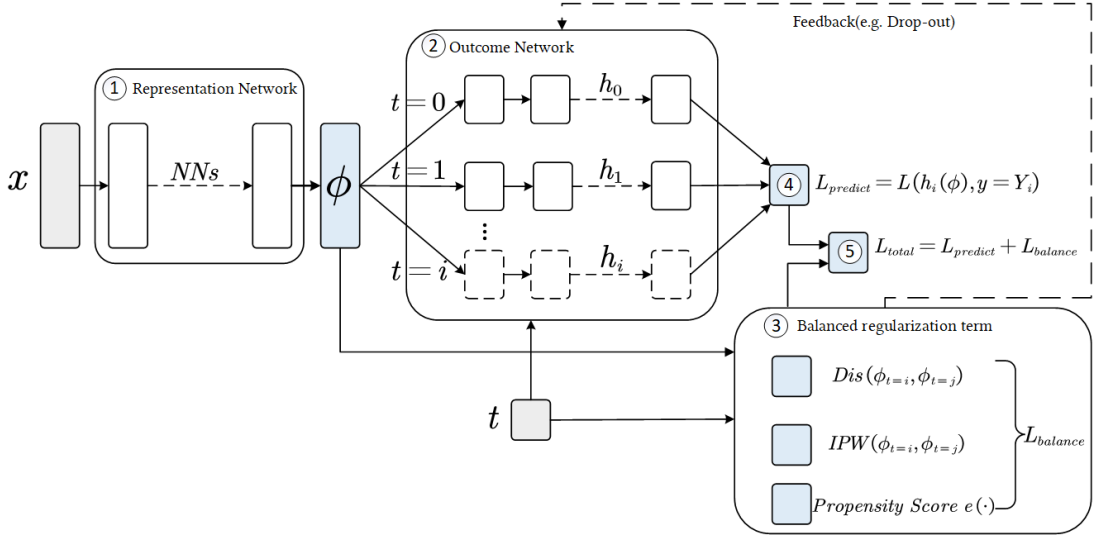

Figure 5: The framework of covariates balance-based causal inference.

To address the problem of the BNN, a new architecture was proposed in the CFR [130], which has two separate heads representing the control group and the treatment group, which share a representation network. This architecture avoids the loss of treatment variable $t$ during network training. In the actual training process, according to the value of $t_i$, each sample is used to update the parameters of the corresponding header. Using the integral probability metric (IPM) [131-132] to measure the distance of control and treated distributions $p(x|t=0)$ and $p(x|t=1)$ was also proposed. In BRNN [43], the MSE is decomposed into bias and variance, and the estimation ability of the multi-head model is compared with that of the single-head model. A new regularization term, PRG, was introduced to assess differences between the treatment and control groups. Here, BWCRF [133] proposed a function called balancing weights to make a trade-off between balance and predictability. It is not a direct balance between groups, instead a weighted balance of representations. In CARW [134], the context-aware weighting scheme that leverages the importance sampling technique based on [130] to better solve the selection bias is integrated. FlexTENet [135] as a new multi-task learning architecture, it is able to adaptively learn what information is shared between potential outcome functions, and improves the estimation of heterogeneous treatment effects by introducing different inductive biases in neural networks. In order to prevent redundancy between shared and private layers and ensure that they encode different information of input features, HTCE [136] proposes an orthogonal regularization loss.

Using the neural network to directly fit the relationship of covariate $X$ to outcome $Y$ can cause many problems. For example, the neural network may use all the variables for $Y$ prediction. In fact, these variables are not needed and should not all be used for estimating causal effects. Covariates can be divided into instrumental variables (only affecting treatment), adjustment variables (only affecting outcome), irrelevant variables (having no effect on either treatment or outcome), and confounders (the cause of both treatment and outcome). When learning representations, SCRNet [137] only balances the representations of confounders. It is then concatenated with the representation of the adjustment variable. This approach reduces computational overhead and increases efficiency in practical applications. However, the division of variables is usually subjective, especially when the true causal graph cannot be obtained. DragonNet [138] is a method for using neural networks to find those covariates that are associated with treatment and only use these variables to predict the outcome. First, a deep neural network is trained to predict $T$, and then the last prediction layer is removed to obtain the representation $\Phi$. Next, similar to the CFR [130], two separate DNNs are used to predict the outcome at $t = 0$ and $t = 1$. Essentially, $\Phi$ stands for the representation related only to $T$, i.e., represents the propensity score. The objective function is:

$$\hat{\theta} = argmin_\theta \hat{R}(\theta; X), where \tag{11}$$

$$\hat{R}(\theta; X) = \frac{1}{n}\sum [(Q^{nn}(t_i, x_i; \theta) - y_i)^2 + \alpha CrossEntropy(g^{nn}(x_i; \theta), t_i)], \tag{12}$$

where $Q^{nn}(\cdot)$ is the DNN to model the outcome and $g^{nn}(\cdot)$ is the DNN to model the propensity score model. $\alpha$ is the hyperparameter to weight the loss term. Simultaneous training to predict propensity scores and outcomes ensures that the features used are treatment-relevant. Compared to CFR [130], DragonNet has part of the predicted propensity score. It makes a clear distinction between the prediction of outcomes and estimation of causal effects because accurate predictions do not mean that causal effects can be accurately estimated. In DragonNet, $g^{nn}(\cdot)$ is used to find the confounding factors in the covariates so that the resulting representation contains only the part related to $X$ (according to the adequacy of the propensity score).

DONUT [139] uses neural networks to learn potential outcomes that are orthogonal to the treatment assignments, combining factual loss and orthogonal regularization loss. It adopts the basic architecture of CFR [130] for the outcome model, and uses logistic regression to estimate the propensity score model. In DCN [140], the architecture of predicting outcome is similar to DragonNet, but it has a separate neural network to model the propensity score. For each head of the predicted outcome DNN, there is a certain probability of dropout [141-142] at each training, and the probability depends on the propensity score. This occurs through dropout, implicitly reflecting the balance effect of the propensity score in the neural network. Deep-treat [143] divides the counterfactual prediction problem into two steps: the first step uses an autoencoder to learn representations that trade off bias and information loss; then the DNN is used on the transformed data to achieve treatment allocation. RSB-Net [144] tackles selection bias with an autoencoder that learns two distinct sets of latent variables: one for selection bias and another for outcome prediction. It employs a loss function regularized by the Pearson Correlation Coefficient to encourage decorrelation between these sets, effectively reducing selection bias and focusing on variables pertinent to individual treatment effect estimation. The balance of covariates helps to eliminate selection bias, but paying too much attention to the balance will affect the counterfactual prediction performance. SITE [145] is a balanced representation learning technique that maintains local similarity while ensuring global balance and preserving predictive power, thus reducing selection bias. It operates in mini-batches, selecting triplet pairs and relying on two main components: PDDM for local similarity preservation and MPDM for balanced latent space distributions. Additionally, a prediction network calculates potential outcome prediction losses. DESCN [146] integrates information about treatment propensity, response, and pseudo treatment effect through an end-to-end multi-task learning approach. This structure not only alleviates the treatment bias and sample imbalance problems, but also the shared network can simultaneously learn the propensity score and the control response with pseudo treatment effects. Causal-StoNet [147] integrates a visible treatment variable within the hidden layer, maintaining computational efficiency. The model adeptly decomposes the joint distribution, capturing conditional probabilities for outcomes, latent factors, treatments, and covariates, thus facilitating precise approximations of both outcome and propensity score functions.

**Multiple treatments**. The previous methods are only applicable to 0-1 treatment case. However, in reality, multiple treatments are often required. For example, when treating patients with multiple chronic diseases or systemic diseases, multiple drugs may need to be used simultaneously. PM [148] extends the approach to multiple treatment settings by using the balanced propensity score to make the match and estimating the counterfactual outcomes using the nearest neighbor. Here, it use the minibatch level, rather than the dataset level, which can reduce variance. RCFR [149] alleviates the bias of BNN [129] when sample sizes are large. It is reweighted according to imbalance terms and variance. NCoRE [150] leverages a conditional neural network with treatment interaction regulators to deduce causal generative processes for various treatment combinations, explicitly modeling cross-interactions through sequential, conditional subnetwork influence. SCP [151] excels in scenarios requiring simultaneous intervention on multiple variables, estimating potential outcomes from single-factor changes using observed data. It expands the dataset through sampling and prediction to enhance balance and sample size, subsequently estimating multi-factor treatment effects on the augmented dataset.

**Continuous treatments**. Estimating the average dose-response function (ADRF) of continuous treatments is an important problem in fields such as healthcare, public policy, and economics. Existing parametric methods are limited in model space, and previous approaches to enhance model expressiveness using neural networks rely on splitting the continuous treatment into chunks and using a separate head for each chunk, which in practice produces discontinuous ADRFs. DRNet [152] incorporates treatment indicators and dosage parameters into the latent representation through a hierarchical structure, using the nearest neighbor similar MISE for model selection without true counterfactuals. It evaluates model performance with metrics like RMSE, dose policy error, and policy error to assess dose-response curve recovery and optimal treatment decisions. VCNet [153] is a variable coefficient neural network that enhances model expressiveness and maintains ADRF continuity, employing B-splines to model treatment level changes continuously. To theoretically support the mitigation of selection bias in a continuous processing setting, ADMIT [154] introduces a theoretical framework to mitigate selection bias in continuous settings, deriving a counterfactual loss upper bound using reweighting and IPM distance. It trains a reweighting network alongside an inference network by minimizing a combined objective function. CMSM [155] is a new marginal sensitivity model of continuous treatment effects, which is based on modeling a set of conditional distributions of potential outcomes and assuming that there is a certain degree of deviation between these distributions, which is bounded by parameters. It is applicable to high-dimensional and large-scale datasets. E2B [156] leverages entropy balancing for end-to-end optimized weights to maximize causal inference accuracy, using pseudo responses and random functions for known response curve estimation.

**Time-varying**. TSD [157] leverages a time series factor model powered by a recurrent neural network to estimate treatment effects amidst numerous hidden confounders. This model employs multi-task outputs to infer latent variables that render treatment assignments conditionally independent, utilizing these variables for causal inference as proxies for the unobserved confounders. SyncTwin [158] addresses irregularly sampled data, particularly for infrequent treatment changes. It employs Seq2Seq to learn temporal covariate representations that capture individualized factors impacting outcomes. An optimization approach identifies weights to construct synthetic twins from control group contributors. The potential outcome is calculated as a weighted average of these contributors' outcomes, with ITE estimated by the difference between the target's outcome and that of their synthetic twin. R-MSN [159] based on the marginal structure modeling framework to learn time-varying treatment responses directly from observational data. It consists of two parts: one is a series of propensity networks used to calculate the treatment probability; the other is a prediction network used to determine the treatment response given a set of planned interventions. DSW [160] harnesses gated recurrent units (GRUs) and attention to discern hidden confounders in time-varying contexts for individual treatment effect estimation. It starts by embedding initial features into a lower-dimensional space and then refines these with a GRU to capture confounder representations. The attention mechanism highlights key historical data, while maximum pooling in the RNN retains early information as time series expand. A final fully connected layer forecasts treatment assignments across time points.

## 5.2 Adversarial Training-based Causal Inference

When there are many covariates and their relationship are particularly complex, sometime the method based on balance is ineffective. This is because the feature representation obtained by the neural network is not reasonable due to the complex relationship among covariates and the limited number of samples. In contrast, the framework based on adversarial training can handle this problem very well. Causal inference methods based on adversarial training mechanisms utilize generative-adversarial networks to implicitly learn counterfactual distributions. Generators are typically used to generate counterfactual results, and then a discriminator is used to determine whether the input data come from the factual distribution or the generate distribution. The basic architecture of adversarial training-based causal inference is shown in Figure 6. It consists of potential outcome generator $G$ and potential outcome discriminator $D$. The potential outcome generator $G$ generates potential outcomes $\tilde{y}$ based on covariates $x$, treatments $t$, and exogenous noise $u$; the generated potential outcomes $\tilde{y}$ and factual outcomes $y$ are then fed into the discriminator $D$, which tries to distinguish which is the factual outcome and which is the generated outcome.

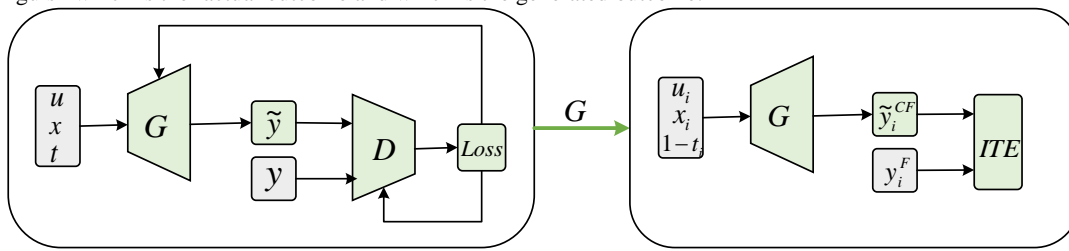

Figure 6: The framework of adversarial training-based causal inference.

GANITE [161] uses the adversarial idea to generate the counterfactual outcomes for a given set of covariates and it can deal with multiple treatment situations. It has two blocks: the counterfactual block was used to impute the "complete" dataset, and the ITE block was used to estimate the causal effect. Due to the advantages of the GANITE framework, the information used for prediction is not lost. At the same time, although there are multiple treatment variables to choose from, only one of them can be selected to match the actual situation. A drawback of GANITE is that it cannot deal with continuous-valued interventions. The SCIGAN [162] addresses this problem very well and provides a theoretical verification of causal estimation under the GAN framework. However, the disadvantage is that it requires thousands of training samples. MatchGAN [163] maps the original samples into the GAN latent space and selects the samples of different categories with the closest distance to enable removal of bias while preserving critical features. CF-Net [164] presents an end-to-end framework for extracting confounder-invariant features that respect the inherent correlation between confounders and outcomes. Its CP module effectively removes confounder influence in feature extraction, showing promise in medical applications. NCMs [165] establish that neural causal models can capture the structural constraints for counterfactual reasoning and propose an algorithm for joint identification and estimation of counterfactual distributions. They introduce a GAN-based approach for robust inference in high-dimensional settings. There is also a class of methods that consider time-related confounding factors, which can help to understand how results change over time. The previously mentioned static-based methods are often not directly applicable to time series scenarios. CRN [166] utilizes a recurrent neural network with adversarial training to generate time-invariant feature representations, effectively neutralizing time-related confounders. This method is well-suited for precision medicine, guiding critical decisions on treatment timing, cessation, and dosage determination.

There are also approaches that combine adversarial and balanced approaches, i.e., using adversarial training to produce covariate-balanced representations [167]. DeepMatch [168] uses adversarial training to balance covariates in such situations. It uses the discriminative discrepancy metric in the context of NNs and requires a few further developments of alternating gradient approaches similar to GAN. ABCEI [167] employs mutual information to gauge the loss during the covariate-to-latent representation transformation, optimizing this measure with a neural network to retain maximum information. It balances group distributions in the latent space through adversarial training, independent of treatment assignment assumptions. CETransformer [169] leverages the Transformer architecture's self-attention mechanism to capture covariate correlations, enhancing feature representation robustness. To address training challenges with limited samples, it uses self-supervised learning with an autoencoder to augment data. Adversarial learning balances group representations, while a two-branch network predicts outcomes based on features and treatment. CBRE [170] introduces a cyclic balanced representation framework, using adversarial training for robust group balance and maintaining data properties through information loops, reducing information loss in latent space. Wasserstein GANs stabilize training, minimizing distribution disparity between groups, with the encoder constrained by a decoder module to preserve data integrity. TransTEE [171] offers a versatile Transformer-based framework for diverse data types, including tabular and sequences, managing various treatment modalities. It selectively utilizes covariates via attention mechanisms, with multi-head cross-attention for adaptive covariate selection.

## 5.3 Proxy Variable-based Causal Inference

For causal inference, there is generally an "unconfounderness" assumption, meaning that there are no unobserved confounders. Most existing methods of estimating causal effects are based on this assumption, include traditional methods or the balance-based methods mentioned earlier. However, in many cases, the assumptions are not satisfied. Once there are unobserved confounders, the previous method will have large bias and even lead to incorrect conclusions. Recent proximal inference provides an alternative approach that can identify causal effects even in the presence of unobserved confounders as long as a sufficiently rich set of proxy variables is measured. It may not be possible to observe all confounders, there is generally a way to measure the proxy variables of the confounders. Proxy variable-based methods utilize proxy variables to separate different types of variables using disentangle representations. Most proxy variables required by the confounders are covered by collecting a large number of observed variables. This is easy to achieve in today's era of big data. Exactly how these proxy variables are used depends on their relationship to the unobserved confounders, treatments, and outcomes.

Deep latent variable techniques can use noisy proxy variables to infer unobserved confounders [172]. CEVAE [173] uses latent variable generative models to discover unobserved confounders from the perspective of maximum likelihood. One of the typical scenarios is shown in Figure 7 (a). This approach requires fairly weak assumptions about the data generation process and the structure of unobserved confounders. It uses the VAE [174-175] architecture and contains both an inference network and a model network. The inference network is equivalent to an encoder, and the model network is equivalent to a decoder. The core of this approach is the use of variational inference to obtain the probability distribution needed to estimate causal effects, $p(X, Z, t, y)$. We use $Z$ to denote the hidden variable, and $X$ is the proxy variable that has no effect on outcome $y$ and treatment $t$. Actually, $Z$ can be seen as the latent variable in VAE. The inference network was used to obtain $q(t|x)$ and $q(z|t,y,x)$. $q(t|x)$ can be seen as the propensity score. The model network was used to obtain $p(t|z)$, $p(x|z)$ and $p(y|t,z)$. Then, the framework is used to model the generation mechanism when unobserved confounders exist:

$$L = \sum_{i=1}^{N} E_{q(z_i|x_i,t_i,y_i)}[logp(x_i, t_i|z_i) + logp(y_i|t_i, z_i) + logp(z_i) - logq(z_i|x_i, t_i, y_i)], \tag{13}$$

$$F_{CEVAE} = L + \sum_{i=1} \big(logq(t_i = t_i^*|x_i^*) + logq(y_i = y_i^*|x_i^*, t_i^*)\big), \tag{14}$$

In practice, to better estimate the distribution of parameters, two terms are added to the lower variable boundary as $F_{CEVAE}$. $y_i^*, x_i^*, t_i^*$ are all the input observed values.

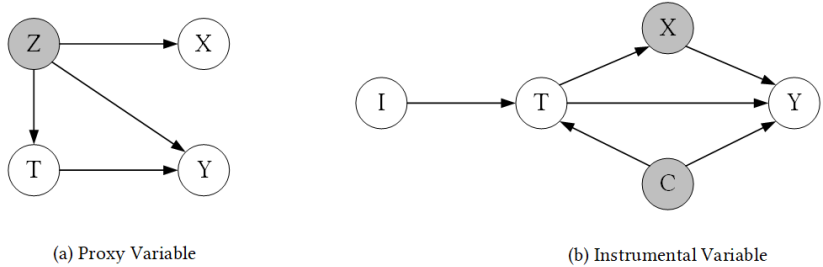

(a) Proxy Variable                    (b) Instrumental Variable

Figure 7: Typical scenarios for proxy variable-based causal inference methods. Gray nodes represent unknown variables or variables that cannot be observed.

In causal effect estimation, an excess of control variables can weaken estimation power, while including non-essential ones may lead to suboptimal nonparametric estimation. High-dimensional scenarios often involve non-confounders that should be omitted to avoid exacerbating bias and variance, thus diminishing estimation precision. The challenge lies in striking a balance: too many variables can inflate bias and variance, while too few might overlook confounders, inducing selection bias. To solve this problem, TEDVAE [176] divides covariates into three categories: confounders, instrumental factors and risk factors. Confounders affect both cause and effect, instrumental factors only affect the cause, and risk factors only affect the effect. The rest of the architecture is similar to CEVAE and can also be used for continuous treatment variables. TVAE [172] improved TEDVAE by combining the method of target learning and maximum likelihood estimation training. Since the causal graph used may vary when making causal inferences, the errors brought by different methods are different, TVAE tends to have a smaller error even though the causal graph is wrong. The purpose of introducing targeted regularization is to make the outcome $y$ and the treatment assignment $t$ as independent as possible. TVAE can be seen as the combination of DragonNet and TEDVAE.

Table 4: Deep causal inference methods.

| Type | Method | Year | Network | Treatment | Hid. Con. | Target |
|------|--------|------|---------|-----------|-----------|--------|
| Balance | BNN [129] | 2016 | MLP | B | N | ITE |
| | CFR [130] | 2016 | MLP | B | N | ITE |
| | DCN [140] | 2017 | MLP | B | N | ITE, ATE |
| | SITE [145] | 2018 | MLP | B | N | ITE |
| | Deep-treat [143] | 2018 | AE, MLP | B | N | ITE |
| | CARW [134] | 2019 | MLP | B | N | ITE |
| | RSB-Net [144] | 2019 | AE | B | N | ITE |
| | DragonNet [138] | 2019 | MLP | B | N | ATE |
| | BRNN [43] | 2020 | MLP | B | N | ITE |
| | BWCRF [133] | 2020 | AE | B | N | CATE |
| | SCRNet [137] | 2020 | MLP | B | N | ITE |
| | DONUT [139] | 2021 | MLP | B | N | ATE |
| | FlexTENet [135] | 2021 | MLP | B | N | CATE |
| | HTCE [136] | 2022 | MLP | B | N | CATE |
| | DESCN [146] | 2022 | MLP | B | N | ITE |
| | Causal-StoNet [147] | 2024 | MLP | B | Y | ATE |
| | PM [148] | 2018 | MLP | M | N | ITE |
| | RCFR [149] | 2018 | MLP | M | N | CATE |
| | NCoRE [150] | 2021 | MLP | M | N | ITE |
| | SCP [151] | 2021 | MLP | M | N | CATE |
| | DRNet [152] | 2019 | MLP | C | N | ITE |
| | E2B [156] | 2021 | MLP | C | N | CATE |
| | VCNet [153] | 2021 | MLP | C | Y | CATE |
| | ADMIT [154] | 2022 | MLP | C | N | CATE |
| | CMSM [155] | 2022 | MLP | C | Y | CATE |
| | R-MSN [159] | 2018 | LSTM | T | N | ATE |
| | TSD [157] | 2019 | RNN | T | Y | ITE |
| | DSW [160] | 2020 | GRU, Attention | T | Y | ITE |
| | SyncTwin [158] | 2021 | LSTM | T | N | ITE |
| Adversarial | GANITE [161] | 2018 | GAN | B | Y | ITE |
| | DeepMatch [168] | 2018 | GAN | B | N | ATE, CATE |
| | ABCEI [167] | 2019 | GAN, AE | B | N | CATE |
| | SCIGAN [162] | 2020 | GAN | C | N | CATE |
| | CF-NET [164] | 2020 | GAN | B | N | Feature |
| | MatchGAN [163] | 2021 | GAN | B | N | ITE |
| | CETransformer [169] | 2021 | GAN, Transformer | B | Y | ITE, ATE |
| | CBRE [170] | 2021 | GAN, AE | B | N | ITE |
| | CRN [166] | 2022 | RNN, MLP | B | Y | ITE |
| | NCM [165] | 2022 | GAN | B, M, C | Y | ITE, ATE |
| | TransTEE [171] | 2022 | Transformer | B, M, C | N | ATE, CATE |
| Proxy | CEVAE [173] | 2017 | VAE | B | Y | ITE, ATE |
| | DeepIV [177] | 2017 | MLP | B, C | Y | ITE |
| | DeR-CFR [178] | 2020 | MLP | B | Y | ITE |
| | TEDVAE [176] | 2020 | VAE | B | Y | ATE, CATE |
| | TVAE [172] | 2020 | VAE | B | Y | ATE, CATE |
| | VSR [179] | 2020 | VAE | B | Y | ITE |
| | CVAE-IV [180] | 2022 | VAE | C | Y | ITE |
| | DMSE [181] | 2022 | MLP | B | Y | ITE, ATE |
| | Credence [182] | 2022 | VAE | B | Y | ATE |
| | CFDiVAE [183] | 2023 | VAE | B | Y | ATE |

Table 4. 'Treatment' indicates the type of treatment. 'B' represents 0-1 variables, 'M' represents multi-variables, 'C' represents continuous variables, and 'T' represents time-varying variables. 'Hid. Con.' indicates if there are hidden confounders.

VSR [179] introduces a variational sample reweighting technique to eliminate confounding bias by decorrelating treatment and confounders, utilizing a VAE for latent treatment representation and deep neural networks for weight estimation. DeR-CFR [178] identifies and balances confounders for treatment effect estimation through decomposition representation, employing reweighting and predictive modeling for counterfactual outcomes. CFDiVAE [183] presents conditional front-gate adjustment (CFD), with theorems ensuring causal effect identifiability, using deep generative models to directly learn CFD variables from data with a provably identifiable VAE. DMSE [181] regards confounders as latent variables, employing conditional independence and variational inference to manage missing data during training and inference for causal effect estimation. Credence[182] assesses causal inference method performance through synthetic data generation, considering user-defined causal effects and biases, based on VAEs that learn data distributions to replicate and evaluate causal inference effectiveness.

Instrumental variable (IV) [184-185] methods look for proxy latent variables for causal inference instead of finding hidden confounders [18]. The IV framework has a long history, especially in economics [186]. The typical scenario using the instrumental variable method is shown in Figure 7 (b). A more powerful instrumental variable approach incorporating deep learning is introduced. DeepIV [177] uses instrumental variables and defines a counterfactual error function to implement neural network-based causal inference in the presence of unobserved confounders. The method can verify the accuracy of the out-of-distribution sample, which is very beneficial and affects the need for hyperparameters for neural network tuning. It is implemented in two steps: the first step is to learn the treatment distribution using a neural network: $\hat{F} = F_\phi(t|x,z)$, where $x$ is the covariate, $t$ is the treatment variable, and $z$ is the instrumental variable. The second step is to use the outcome network to predict the counterfactual outcomes. The objective function is:

$$L(D; \theta) = |D|^{-1} \sum_i \left( y_i - \int h_\theta(t, x_i) d\hat{F}_\phi(t|x_i, z_i) \right)^2,$$

(15)

where $h$ is the prediction function, $\hat{F}_\phi$ is the treatment distribution obtained from the first step, and $D$ is the dataset. Confounded instruments violate a key condition in the instrumental variable assumption—that unmeasured confounders and instrumental variables are conditionally independent given observed covariates. To deal with this situation, CVAE-IV [180] proposed an alternative hidden confounder by considering conditional independence to meet the strong ignorability criterion, thereby allowing unbiased estimation of ITE. Table 4 shows all the mentioned deep causal inference methods.

## 6 NEW FRONTIERS OF CAUSAL LEARNING

In addition to the deep causal learning methods previously described, there are several cutting-edge causal learning areas that warrant close attention. These fields have made rapid progress recently and have achieved outstanding performance in some tasks.

### 6.1 Supervised Causal learning

Existing causal discovery mainly relies on artificial prior knowledge to determine causal relationships, or uses non-causal supervisory information to learn causal relationships, rather than learning from data with causal supervisory information. For example, the constraint-based method uses conditional independence tests to determine whether there is a causal relationship between variable pairs, which causes a large amount of information in the data that may be related to causality to be omitted, resulting in low accuracy and efficiency; the score-based method uses the minimum reconstruction loss to learn causal relationships, resulting in some correlations being used to determine whether causal relationships exist or not.

Causal supervisory information has rarely been used in previous causal discovery methods because the causal graph corresponding to real observation data is very difficult to obtain. In the field of computer vision, obtaining supervisory information may be labeling images; in the field of natural language processing, obtaining supervisory information may be making emotional judgments on sentences; and in the field of causal discovery, obtaining "causal graphs" as supervisory information requires a full understanding of the generation mechanism behind the variables involved, so causal graph supervisory information has not been taken seriously in the field of causal discovery.

Recently, many studies have introduced the paradigm of supervised learning into the field of causal discovery. This paradigm is called supervised causal learning. Supervised causal learning synthesizes a large amount of data with real causal graphs to train a model that can identify causal relationships, allowing the model to learn what the real causal relationship is from a large amount of data. ML4S [187] and ML4C [188] use supervised learning to learn skeleton and v-structure respectively. Lorch et al. [189] proposed the amortized variational inference model, which uses the inductive bias of a specific domain represented by the graph scoring function, uses the synthesized causal graph and the corresponding sampled data for training, and uses continuous optimization to obtain the inference model. The model architecture uses the alternating attention mechanism [190] to permute and equivariant the sample dimension and variable dimension of the dataset respectively, and proves that the supervised causal learning paradigm is generalizable and feasible on independent and identically distributed data. Ke et al. [191] proposed the CSIvA model, which also uses synthetic data generated by different causal graphs to train the model, and also learns the mapping from data to graph structure based on the alternating attention mechanism. At the same time, it extends the supervised learning paradigm to also use intervention data, thereby achieving greater flexibility.

Although supervised causal learning methods can define causality more "objectively" due to their data-driven form, they need to additionally address the generalization challenge that all supervised methods face, namely the distribution bias between training samples and test samples. This will be the main research direction of supervised causal learning in the future.

## 6.2 End to End Causal learning

Existing causal discovery and causal inference methods often develop independently. In order to obtain the causal effect between variables, it is necessary to have some understanding of the causal relationship between the variables, that is, it is based on the causal graph that has been obtained, which leads to the performance of the causal discovery task affecting the performance of the causal inference task. With the in-depth integration of neural networks and causal learning, some end-to-end deep causal learning algorithms have recently been developed to achieve simultaneous causal discovery and causal inference based on observational or intervention data.

By fully combining the deep mechanism with SCM, DSCM [192] can use exogenous variables for counterfactual inference via variational inference. Three types of mechanisms combined with SCM have been discussed: explicit likelihood, amortized explicit likelihood, and amortized implicit likelihood. These three mechanisms may need variational inference and normalizing flows to model. DSCM can also finish the three steps of counterfactual inference depicted by Pearl [1], which are abduction, action, and prediction. Toth et al. [127] proposed a new framework called Active Bayesian Causal Inference (ABCI), which aims to integrate causal discovery and causal reasoning through active learning. Compared with the traditional two-stage method (first inferring the causal graph and then estimating the causal effect of the intervention), ABCI is more efficient, especially in terms of actively collected intervention data. It mainly considering hard interventions, and providing corresponding intervention probability calculation methods. ABCI mainly considers nonlinear additive Gaussian noise models. From a Bayesian perspective, causal queries are regarded as potential quantities and posterior inferences are performed, while uncertainty is considered in the process. DECI [193] uses additive noise model to describe the causal relationship between variables, where each variable is a function of its parent variable and exogenous noise. Based on autoregressive flow, variational inference is used to learn the posterior distribution of the causal graph, which is capable of learning complex nonlinear relationships between variables and non-Gaussian exogenous noise distributions. The learned generative model is used to evaluate the expectations under the intervention distribution, thereby estimating ATE and CATE. At the same time, in order to adapt to real data, the model is extended to support mixed type (continuous and discrete) data and the processing of missing values. Zhang et al. [194] explored the duality of causal inference and attention mechanisms in Transformer-based architectures, demonstrated the primal-dual connection between optimal covariate balancing and self-attention, and proved that trained self-attention can find the optimal balancing weights for a given dataset under appropriate self-supervised losses. They propose CInA to use multiple unlabeled datasets for self-supervised causal learning and achieve zero-shot causal inference on unseen tasks and new data. The model uses self-attention as its last layer and is trained to learn the optimal covariate balancing weights. Once the model is trained on multiple data sources, it can perform zero-shot causal inference by simply extracting the key-value tensor of the last layer of the model.

## 6.3 Large Language Models for Causal Learning

With the recent development of large language models (LLMs), iterative progress has been promoted in various scientific fields. Due to its powerful representation ability, understanding of complex data structures, and rich world knowledge, LLMs can provide assistance in many aspects of causal learning.

There are now many studies investigating the performance of LLMs on various causal tasks. Kıcıman et al. [195] conducted an in-depth analysis of LLMs, distinguishing different types of causal reasoning tasks. They found that algorithms based on GPT-3.5 and 4 achieved new state-of-the-art accuracy on multiple causal benchmarks. Specifically, they surpassed existing algorithms on paired causal discovery tasks, counterfactual reasoning tasks (92% accuracy, an increase of 20 percentage points), and actual causality (86% accuracy in determining necessary and sufficient causes in cases). Zhang et al. [196] argue that current LLMs can combine existing causal knowledge to answer causal questions like domain experts, but they cannot yet provide satisfactory, precise answers for discovering new knowledge or making high-stakes decision-making tasks. The possibility of enabling explicit and implicit causal modules and deep causal-aware LLMs is discussed, which will not only enable LLMs to answer more different types of causal questions, but also make LLMs more trustworthy and efficient in general. Cai et al. [197] demonstrated through experiments that the causal reasoning ability of LLMs depends on the context and domain-specific knowledge provided. When appropriate knowledge is provided, LLMs exhibit causal reasoning capabilities consistent with human logic and common sense; in the absence of knowledge, LLMs can still use existing numerical data to perform a certain degree of causal reasoning, but there are certain limitations.

***LLMs for causal discovery.*** Long et al. [198] explored how to use the expert knowledge of LLM to improve data-driven causal discovery, especially how to use the expert knowledge to determine the direction of causal relationships between variables when experts may provide wrong information. Kıcıman et al. [195] found that LLMs outperformed existing data-driven causal discovery algorithms on paired causal tasks. LLMs are able to infer causal relationships by analyzing the metadata of variables (such as natural language descriptions of variable names and problem contexts), which is a different approach from previous causal discovery algorithms that rely on data values. In full-graph causal discovery tasks, LLMs are able to handle more complex data sets and are comparable to domain experts and deep learning-based methods. Thomas [199] proposed a method for causal discovery using LLMs based on breadth-first search, which can improve the performance of causal discovery tasks under different causal graph sizes by only using linear query times. Ahmed et al. [200] proposed a framework that combines the metadata-based reasoning capabilities of large language models with data-driven modeling of deep structural causal models for causal discovery tasks. First, a causal graph is generated using the world knowledge of the LLM, and a data-driven approach is used to calculate the model's fit. Long et al. [201] used four causal graphs representing known medical literature as ground truth. For each causal graph, a loop was traversed over each ordered variable pair and GPT-3 was asked to score both directions separately. The experiment found that the accuracy of GPT-3 in confirming the direction of two variables depends on the language used to describe the relationship. At the same time, the update of LLMs and the data they are trained on lag behind the availability of new medical literature and may not be suitable for constructing DAGs for new diseases. Tu et al. [202] assessed ChatGPT's causal discovery capabilities in neuropathic pain diagnosis, testing it with 50 true and 50 false causal relationship pairs from a dataset. They found that while ChatGPT could provide plausible responses, it lacked a true grasp of the subject matter, such as the human nervous system. The model exhibited performance variability, offering inconsistent answers to identical questions over time. It struggles with comprehending new concepts beyond its training data, highlighting a need for enhanced consistency and reliability.

*LLMs for causal inference.* CURE [203] is based on the Transformer model and uses a pre-training and fine-tuning approach to convert structured longitudinal patient data into sequence inputs by temporally aligning all observed covariates. It is pre-trained on large-scale unlabeled patient data, uses unsupervised learning objectives to generate contextualized patient information representations, and fine-tunes the pre-trained model on downstream labeled treatment and outcome data. Jin et al. [204] proposed a new task CORR2CAUSE, created the first benchmark dataset for testing the pure causal reasoning ability of LLMs, containing more than 200K samples, and used this dataset to evaluate seventeen existing LLMs. The CORR2CAUSE task requires LLMs to accept a set of correlation statements and determine the causal relationship between variables. Experiments show that existing LLMs perform poorly on causal inference tasks, close to random levels. Even though fine-tuning can improve performance to a certain extent, these models still cannot generalize well to out-of-distribution settings. Jin et al. [205] created a benchmark dataset called CLADDER to explore the causal inference ability of LLMs, which contains 10,000 samples. These samples are based on a series of causal graphs and queries (including associations, interventions, and counterfactuals), and a prophetic causal inference engine generates symbolic questions and real answers, which are then translated into natural language. The performance of multiple LLMs on the CLADDER dataset was evaluated, and a customized thought chain prompt strategy called CAUSALCOT was introduced and evaluated. Through experiments on the CLADDER dataset, the authors found that the most advanced models face significant challenges in causal reasoning tasks.

## 7  CONCLUSION AND FUTURE DIRECTIONS

Deep learning models show great advantages in various fields, and an increasing number of researchers are trying to combine deep learning with causal learning. It is hoped that the powerful representation and learning abilities of neural networks can enhance existing causal learning methods in various ways. This article reviews three aspects of the improvements that deep learning brings to causal learning: causal representation learning, deep causal discovery, and deep causal inference. Causal representation learning uses neural networks to learn the representation of causal variables, especially in the case of unstructured high-dimension data. We reviewed deep causal discovery methods based on neural networks based on different data types. Then, we reviewed causal inference methods using deep learning from three perspectives: covariate balance-based, adversarial training-based, and proxy variable-based methods. Finally, we showcase some of the most advanced areas of causal learning.

Although deep learning has brought about significant changes to causal learning, there are still many problems that must be addressed. In this section, we pose these questions and included a brief discussion, with the hope of offering researchers some future directions.

***Scarcity of causal knowledge and utilizing strong assumptions.*** Although many of us are devoted to the field of causal learning, as researchers, we must constantly reflect on whether what we think of as causality is truly causality and whether there is a more suitable form of studying causality than the causal graph or potential outcomes. Because of this lack of causal knowledge, we rely heavily on untestable assumptions when studying causality. In the causal discovery field, the correctness of the causal Markov condition assumption and faithfulness assumption still needs to be fully verified [206]. SUTVA, unconfounderness, and positivity assumptions are required when making causal inference. The conditions under which these assumptions apply and fail need to be fully studied (e.g., due to social connections, one person's treatment outcomes may affect another person through social activities.). Most existing methods and applications of causal inference are based on the assumption of directed acyclic graphs [207], but in reality, there may be causal feedback between variables, leading to the emergence of cyclic graphs [105]. In addition, different methods are based on different assumptions about the distribution of noise [31, 208]. How to reasonably relax the assumptions while ensuring the accuracy is a very challenging problem. Although these assumptions are convenient, they also bring many risks, and care must be taken when using them.

***Lack of casual benchmarks and suitable metrics.*** Although there are some commonly used datasets in the fields of causal discovery and causal inference [27, 47], the scale and number of these datasets make it difficult to convince people of the model's capabilities. Moreover, most of the existing datasets are synthetic data, so it is crucial to collect more real datasets and develop high-quality simulators. At the same time, the metrics used to estimate causal effects on different datasets remain inconsistent. For example, $\epsilon PEHE$ used in IHDP is only suitable for binary treatment variables, while the metric $R_{pol}$ used in the Jobs dataset is highly targeted and does not have universality. Rich and diverse causal "loss functions" similar to the current stage of deep learning [209-211] would be very helpful for the rapid improvement of the performance of deep causal learning algorithms.

***Combine data-driven with knowledge-driven.*** Current empirical studies on LLMs have shown that they currently rely mainly on the world knowledge behind the metadata to complete causal learning related tasks. It is a very important direction to further verify whether LLMs have true causal inference capabilities and their degree of dependence on knowledge and data respectively. At the same time, most of the current causal discovery and causal inference methods are data-driven. How to effectively combine these methods with knowledge-driven LLMs depth is also worthy of study.

***Causality for deep learning.*** In this article, we mainly discussed the changes that deep learning brought to causal learning. At the same time, causal learning is profoundly changing the field of deep learning. Many studies have focused on how causality can help deep learning address long-standing issues such as interpretability [37-38, 40, 212-213], generalization [214-216], robustness [217-219], and fairness [220-222].

## ACKNOWLEDGMENTS

We thank Xingwei Zhang and Songran Bai for their precious suggestions and comments on this work. We also thank Haitao Huang and Gang Zhou for invaluable discussion. This work is supported by the Ministry of Science and Technology of China under Grant No. 2020AAA0108401，and the Natural Science Foundation of China under Grant Nos. 72225011 and 71621002.